%% file: main.tex
\documentclass[letterpaper,10pt,conference]{ieeeconf}
\IEEEoverridecommandlockouts
\usepackage{graphicx}
\usepackage{microtype}
\usepackage{needspace}
\newcommand{\panellabel}[1]{%
  \parbox[t]{.14285\textwidth}{\centering\footnotesize\strut#1\strut\par}%
}
\usepackage{amsmath}
\usepackage{amssymb}
\usepackage{booktabs}
\usepackage{multirow}
\usepackage{array}
\usepackage[nocompress]{cite}
\usepackage[hyphens]{url}
\usepackage{placeins}
\usepackage{etoolbox}
\makeatletter
\patchcmd{\@makecaption}
  {\begin{center}{\footnotesize #1}\\{\footnotesize\scshape #2}\end{center}}
  {\par\begingroup\centering\footnotesize #1:\enspace {\scshape #2}\par\endgroup}
  {}
  {\PackageWarning{compactlayout}{Table caption patch failed}}
\renewcommand{\section}{\@startsection{section}{1}{\z@}%
  {1.2ex plus .4ex minus .2ex}{.5ex plus .2ex}%
  {\normalfont\normalsize\centering\scshape}}
\renewcommand{\subsection}{\@startsection{subsection}{2}{\z@}%
  {1.2ex plus .4ex minus .2ex}{.5ex plus .2ex}%
  {\normalfont\normalsize\itshape}}
\let\NAT@parse\undefined

\makeatother

\usepackage[colorlinks=true,citecolor=blue,linkcolor=black,urlcolor=black]{hyperref}

\title{\LARGE\bf NeuralSRender: G-Buffer Guided Video Diffusion for Sim-to-Real Rendering}
\author{Yipeng Lu, Youqi Liao, Zhenyu Wang, Yuning Peng, and Zhen Dong%
\thanks{All authors are with Wuhan University, Wuhan, China
(e-mail: \mbox{\href{mailto:dongzhenwhu@whu.edu.cn}{dongzhenwhu@whu.edu.cn}}).}}
\begin{document}
\maketitle
\thispagestyle{empty}
\pagestyle{empty}

\input{Parts/00_abstruct}

\input{Parts/01_introduction}
\input{Parts/02_Related_Work}

\input{Parts/03_Dataset}

\input{Parts/04_Method}
\input{Parts/05_Experiments}
% Allow the conclusion to fill space while pending figures are placed.
\input{Parts/06_Conclusion}

\bibliographystyle{IEEEtran}
\bibliography{references}
\end{document}

%% file: Parts/00_abstruct.tex
\begin{abstract}

  % High-quality grounded data remain a critical bottleneck for foundation models in physical-world applications, such as autonomous driving and embodied AI. Simulators offer a scalable alternative to costly real-world data collection, but the domain gap between rendered and real images often limits the effectiveness of sim-to-real methods.
   % NeuralSRender is highly parameter-efficient, adding only 1.45\% parameter overhead. It can be fully trained on consumer-grade GPUs with 24 GB of memory while maintaining competitive inference speed. We also introduce GenReal, a large-scale autonomous driving dataset that provides synthetic and real-world videos with aligned G-buffers.
   % In this paper, we propose NeuralSRender, a sim-to-real video translation framework that combines graphics priors (G-buffers) with the generative capabilities of video diffusion models.
   % Extensive experiments show that NeuralSRender achieves the best visual realism while preserving competitive geometric consistency among baselines. Its generated data improve zero-shot object detection on nuImages by 15.7\% and semantic segmentation on Cityscapes by 25.69\%. 
  The scarcity of high-quality annotated visual data remains a critical bottleneck for autonomous-driving perception. Simulators provide training data at scale, but the limited realism of synthetic videos compromises perception performance in real-world applications.
To address this limitation, we propose NeuralSRender, a sim-to-real video translation framework that enhances the photorealism of synthetic videos while maintaining geometric consistency with the source scenes.
Built on a pretrained video diffusion transformer (DiT), NeuralSRender uses a lightweight projector to inject geometric and material priors from G-buffers, and LoRA enables parameter-efficient adaptation.
The model is trained in two stages: it first learns G-buffer conditioning from synthetic videos with accurate G-buffers, then adapts to real-world appearance using real videos with estimated G-buffers. During the second stage, an additional LoRA module is trained while the projector and first-stage LoRA remain frozen to reduce interference from estimation noise.
NeuralSRender introduces only 1.45\% parameter overhead and can be trained on consumer-grade GPUs with 24 GB of memory.
To support training and evaluation, we introduce GenReal, a large-scale driving video dataset comprising synthetic and real-world videos with corresponding G-buffers.
Experiments demonstrate improved photorealism with competitive geometric consistency. Training perception models on our translated videos instead of raw simulator renders improves detection $\mathrm{AP}_{50}$ on nuImages from 35.1 to 40.6 and segmentation mIoU on Cityscapes from 32.4 to 40.8.
Our code, data, and model weights are available at: {\hypersetup{urlcolor=blue}\url{https://github.com/WHU-USI3DV/NeuralSRender}}.

\end{abstract}

%% file: Parts/01_introduction.tex
\section{Introduction}
\label{sec:intro}

\begin{figure*}[!t]
  \centering
  \includegraphics[width=\textwidth]{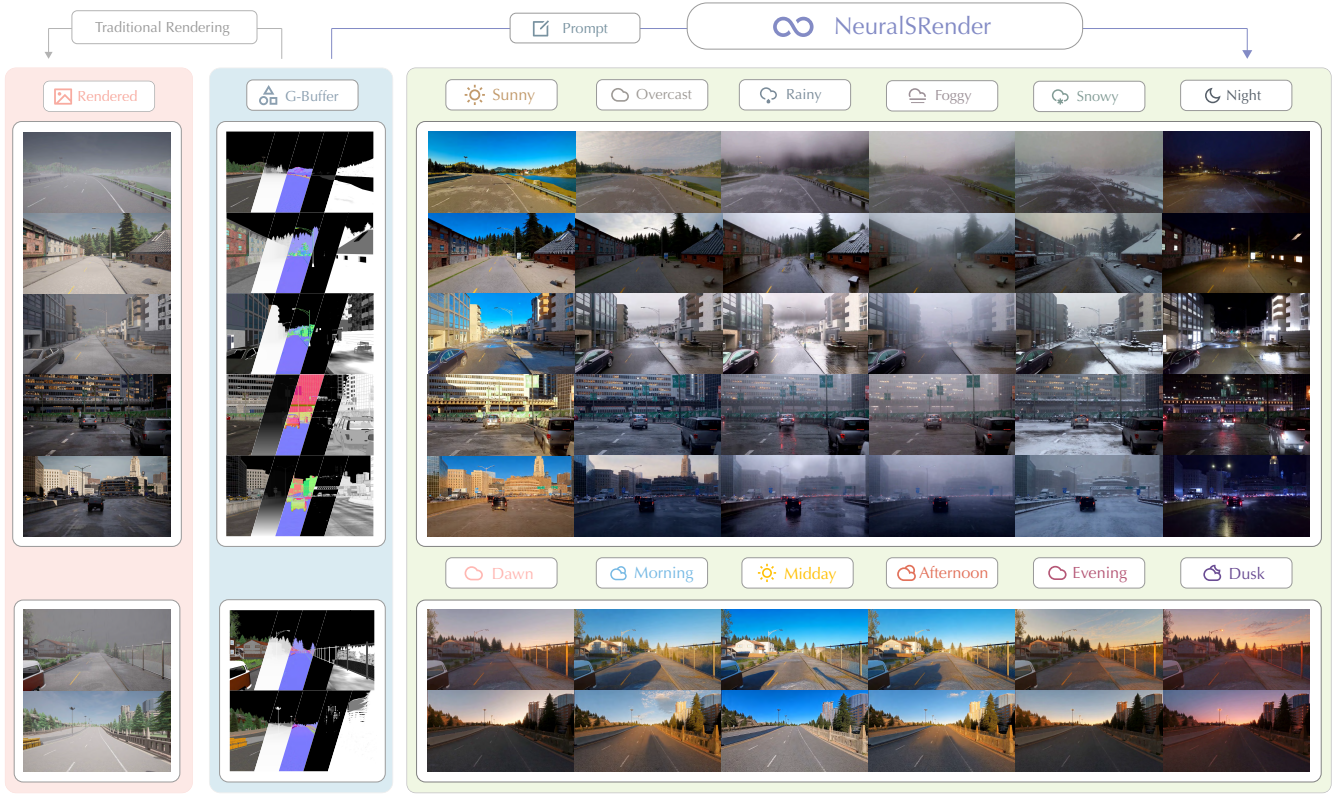}

  \caption{Photorealistic sim-to-real synthesis with the proposed NeuralSRender. Our approach leverages video generation priors to synthesize photorealistic driving videos from G-buffers, reducing reliance on high-fidelity textures and materials while enabling text-guided changes in weather and time of day.}  \label{fig:tesear}

\end{figure*}

Training models for autonomous driving~\cite{gao2023magicdrive,hu2023planning,hwang2025emma,gao2025magicdrive-v2,liao2025diffusiondrive} requires large-scale, diverse driving data and task-specific supervision.
 However, collecting and annotating real-world driving data at scale is costly and labor-intensive. Moreover, routine data collection struggles to capture rare but safety-critical scenarios, leaving gaps in training-data coverage.
% For these physical-world applications, web-scale visual data alone are insufficient, as accurate geometry and task-relevant annotations are often unavailable. Collecting suitable real-world data is costly and provides limited coverage of corner-case scenarios, making data scarcity a key barrier to scaling up.

Simulators provide a scalable alternative to real-world data collection~\cite{Dosovitskiy17carla,gaidon2016virtualkitti,cabon2020vkitti2,shah2017airsim}, offering automatic annotations and controlled generation of rare and safety-critical driving scenarios. However, their rendered images and videos often differ substantially in appearance from real-world data, which can limit the real-world generalization of models trained on synthetic data.~\cite{kendall2019learning}.
This appearance gap reflects practical limitations in both asset creation and rendering. First, creating realistic textures and materials is costly and difficult to scale across diverse scenes. Meanwhile, real-time rendering engines often rely on simplified models of light transport to meet computational budgets, limiting their ability to reproduce complex lighting effects. These limitations motivate methods to improve photorealism in simulated data while preserving scene geometry and annotation consistency.

Recent advances in controllable video generation and neural rendering offer promising avenues for sim-to-real translation.
Controllable video generation methods use semantic, structural, or motion cues to guide generation~\cite{cosmos_cookbook_control_modalities_2025,gao2025magicdrive-v2,gu2025das,ren2025cosmos}.
Although these cues help constrain scene layout and motion, they do not generally provide a joint, explicit description of geometry and material properties.
Neural rendering methods incorporate richer scene representations, including geometric and material attributes, to support controllable appearance synthesis~\cite{DiffusionRenderer,liu2025tclight,he2025unirelight}.
However, leveraging these attributes for sim-to-real translation presents a challenge in training: synthetic videos provide accurate simulator-derived attributes but limited appearance realism, whereas real videos offer realistic appearance but typically rely on noisy estimates of the corresponding attributes.
The key challenge is therefore to learn reliable geometric and material conditioning while adapting to real-world appearance, improving photorealism without compromising scene structure and annotation consistency.

To address this challenge, we propose NeuralSRender, a parameter-efficient framework for G-buffer-guided sim-to-real video translation. As illustrated in Fig.~\ref{fig:tesear}, NeuralSRender synthesizes
realistic driving scenes under diverse text-specified weather and
time-of-day conditions while preserving the overall scene layout.
A lightweight projector integrates dense geometric and material features into a pretrained video diffusion model.
Our two-stage training strategy separates reliable conditioning learning from real-world appearance adaptation.
We first jointly train the projector and Stage-I LoRA modules on synthetic videos with accurate G-buffers.
We then freeze both and train additional LoRA modules on real videos with estimated G-buffers~\cite{DiffusionRenderer} for appearance adaptation.
The framework introduces only a 1.45\% parameter overhead relative to the backbone.
We also introduce GenReal, a large-scale autonomous-driving dataset spanning two synthetic domains and one real-world domain.
GenReal brings together driving videos, simulator-derived or estimated G-buffer representations, and descriptive text prompts, providing a data resource for learning geometric and material conditioning and systematically evaluating sim-to-real video translation across domains.
Experiments demonstrate that NeuralSRender achieves superior visual realism among the evaluated baselines while maintaining competitive geometric consistency.
Models trained on our translated synthetic data outperform those trained on other synthetic data in object detection and semantic segmentation on real-world benchmarks. 
Our main contributions are summarized as follows:

\begin{itemize}
\clubpenalty=10000
\widowpenalty=10000
   \item We propose \textbf{NeuralSRender}, a parameter-efficient framework for sim-to-real video translation.
It combines dense geometric and material conditioning with pretrained video generative priors, introducing only a 1.45\% parameter overhead relative to the backbone.
    \item We develop a two-stage training strategy to improve photorealism while preserving reliable G-buffer control. It first learns conditioning from accurate synthetic G-buffers, then adapts appearance using real videos while freezing the learned conditioning components to mitigate interference from noisy attribute estimates.
   \item We construct \textbf{GenReal}, a large-scale driving video dataset spanning two synthetic domains and one real-world domain, with G-buffer annotations and text prompts for training and cross-domain evaluation.
\end{itemize}

%% file: Parts/02_Related_Work.tex
\section{Related Work}
\label{sec:related_work}

\paragraph{\textbf{Controllable Video Generation.}}

Controllable video generation aims to synthesize videos under spatial or semantic conditions such as text, pose, depth, and edges. Methods including ControlNet~\cite{zhang2023controlnet}, T2I-Adapter~\cite{mou2024t2iadapter}, VideoComposer~\cite{wang2023videocomposer}, SparseCtrl~\cite{guo2024sparsectrl}, and Ctrl-Adapter~\cite{lin2024ctrl} introduce trainable branches or lightweight conditioning modules to inject external controls. Recent video frameworks such as VACE~\cite{jiang2025vace} and CogVideoX~\cite{yang2025cogvideox} extend this paradigm with temporal modeling and more flexible conditioning. However, most existing methods rely on sparse or high-level control signals and do not explicitly model the dense geometry and material interactions required for sim-to-real rendering. In contrast, NeuralSRender conditions a pretrained video diffusion model on multi-attribute G-buffers, providing aligned geometric and material guidance for realistic video translation.

\input{Parts/021_data_table}

\paragraph{\textbf{Sim-to-Real Translation.}}
Sim-to-real translation aims to reduce the visual gap between simulated and real-world data. Early image-based methods, such as RCAN~\cite{james2019sim}, transform randomized simulator outputs into a canonical domain. A recent diffusion-based method~\cite{samak2025sim2real} further improves visual realism using text prompts or reference images. However, image-level translation does not model temporal consistency. Video-to-video synthesis~\cite{wang2018video} generates temporally coherent sequences from semantic maps, but does not explicitly exploit the dense geometry and material information available in simulators. In this work, NeuralSRender uses G-buffers as a shared structural representation across synthetic and real domains, providing dense guidance for sim-to-real video translation.

\paragraph{\textbf{Neural Rendering.}}

Recent studies have used diffusion priors for image and video relighting without explicit 3D reconstruction. IC-Light~\cite{zhang2025iclight} models light transport consistency in the latent space, while Neural Gaffer~\cite{jin2024neural_gaffer} uses synthetic data to handle complex illumination and reflective materials. For video relighting, Light-A-Video~\cite{Zhou_2025_lightavideo} improves cross-frame consistency, and TC-Light~\cite{liu2025tclight} preserves dynamic geometry while adjusting global illumination. However, these methods mainly focus on lighting changes. They do not fully address the broader appearance adaptation required for sim-to-real translation, including changes in textures, materials, and weather. DiffusionRenderer~\cite{DiffusionRenderer} extracts G-buffers from videos and uses them as structural conditions for forward video diffusion. However, it mainly targets object-centric relighting and does not use text guidance. This limits flexible control over global appearance and large domain shifts. In contrast, NeuralSRender combines dense G-buffer conditions with text guidance and a two-stage adaptation strategy for sim-to-real video translation in complex outdoor scenes.

%% file: Parts/021_data_table.tex
\begin{table*}[t]
  \centering
  \caption{Detailed statistics of the GenReal dataset.}
  \label{tab:data}

  \footnotesize

  \begin{tabular*}{0.8\textwidth}{
    @{\extracolsep{\fill}}llllcccc@{}
  }
    \toprule
    Domain
    & Source
    & Subset
    & Split
    & Count
    & Resolution
    & Frames
    & G-buffer \\
    \midrule

    \multirow{3}{*}{Synthetic}
    & \multirow{2}{*}{CityDataset}
    & \multirow{2}{*}{GenReal-City}
    & Train
    & 1,795
    & $832 \times 480$
    & 81
    & Accurate \\

    &
    &
    & Test
    & 140
    & $1280 \times 704$
    & 81
    & Accurate \\

    \cmidrule(lr){2-8}

    & CARLA~\cite{Dosovitskiy17carla}
    & GenReal-CARLA
    & Test
    & 1,143
    & $1280 \times 704$
    & 81
    & Accurate \\

    \midrule

    \multirow{2}{*}{Real}
    & \multirow{2}{*}{OpenDV~\cite{yang2024genad}}
    & \multirow{2}{*}{GenReal-DV}
    & Train
    & 5,819
    & $832 \times 480$
    & 81
    & Estimated \\

    &
    &
    & Reference
    & 7,903
    & $1280 \times 720$
    & 81
    & None \\

    \bottomrule
  \end{tabular*}

\end{table*}

%% file: Parts/03_Dataset.tex
\section{GenReal Dataset}

Existing datasets rarely provide both realistic driving videos and dense rendering attributes.
Real-world datasets~\cite{nuscenes2019,sun2020scalability,yang2024genad} capture realistic visual appearances but lack G-buffer annotations, whereas simulators~\cite{Dosovitskiy17carla,shah2017airsim} provide accurate G-buffers but suffer from a large visual domain gap.
To support sim-to-real video translation, we construct GenReal, which consists of two synthetic subsets, GenReal-City and GenReal-CARLA, and one real-world subset, GenReal-DV. Each annotated sample contains an RGB video, its corresponding basecolor, depth, normal, metallic, and roughness sequences, and a descriptive text prompt.
Synthetic G-buffers are directly exported from the rendering pipelines, while real-world G-buffers are estimated using the pretrained inverse-rendering model DiffusionRenderer~\cite{DiffusionRenderer}.
Table~\ref{tab:data} summarizes the dataset statistics.

\paragraph{Synthetic Data Construction.}
GenReal-City is collected in the Unreal Engine 5 CitySample environment with the MatrixCity plugin~\cite{li2023matrixcity}. We use the built-in traffic system to generate moving vehicles and record driving trajectories. These trajectories are then rendered with Movie Render Queue to obtain RGB frames and their corresponding G-buffers.
GenReal-CARLA is collected from multiple CARLA maps using the built-in autonomous driving agent. We record rendered RGB sequences together with the corresponding G-buffer sequences.
The two synthetic sets differ in scene layouts, asset quality, and rendering styles, which increases the diversity of the synthetic data.

\paragraph{Real-World Data Construction.}
GenReal-DV is constructed from real-world driving videos in the OpenDV dataset.
We sample video clips from the source videos and use the pretrained DiffusionRenderer~\cite{DiffusionRenderer} to estimate basecolor, depth, normal, metallic, and roughness maps.
We manually remove clips with severe motion blur, ghosting, extreme exposure, or lens occlusion, while retaining clips with mild real-world artifacts to preserve the diversity of natural driving scenes. Unlike the accurate G-buffers exported by simulators, the estimated attributes in GenReal-DV contain noise, making the dataset suitable for studying robust adaptation from synthetic to real-world domains.

\paragraph{Text Prompt Construction.}
Each video is paired with a text prompt generated through a hybrid human--VLM annotation pipeline.
Human annotators label low-level visual attributes that are difficult for VLMs to estimate reliably, including weather, time of day, exposure, color style, camera configuration, sunlight direction, and imaging artifacts.
Qwen3-VL~\cite{bai2025qwen3} is used to extract high-level semantic information, such as scene type, object layout, road conditions, and background content.
% Detailed annotation options, quality-control procedures, and prompt templates are provided in the supplementary material.

%% file: Parts/04_Method.tex
\section{Method}
\label{sec:method}

\begin{figure*}[!t]
  \centering
  \includegraphics[width=\textwidth]{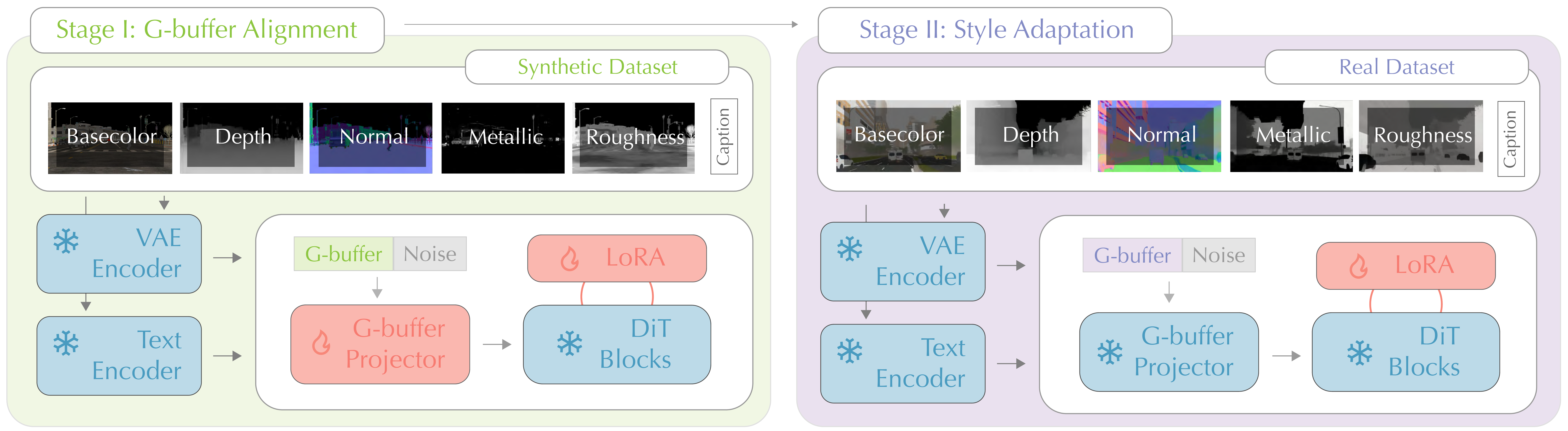}

\caption{
Overview of NeuralSRender. In Stage I, the G-buffer projector and LoRA modules are trained on synthetic videos with accurate G-buffers. In Stage II, the model is adapted to real-world videos using estimated G-buffers. In Stage II, the projector is frozen to protect the conditioning mechanism learned from synthetic data, while only the LoRA modules are optimized.
}

\label{fig:pipeline}

\end{figure*}

NeuralSRender generates photorealistic driving videos from G-buffers and text prompts, as illustrated in Fig.~\ref{fig:pipeline}.
A lightweight projector integrates geometric and material features into a pretrained video diffusion model, with the backbone remaining frozen during LoRA adaptation.
Our two-stage training strategy first learns conditioning from accurate synthetic G-buffers, then adapts to real-world appearance using real videos while keeping the learned conditioning components fixed.

\subsection{G-buffer Condition Fusion}\label{sec:fusion}

G-buffers provide dense geometry and material information for rendered videos.
Depth and normals represent scene geometry, while basecolor, metallicity, and roughness characterize material appearance.
For a video of \(T\) frames, let
$
g^{(i)} = \left\{g_k^{(i)}\right\}_{k=1}^{T}
$
denote the temporal sequence of the \(i\)-th G-buffer attribute, and let
$
\mathbf{g} = \{g^{(i)} \}_{i=1}^n
$
denote the complete set of G-buffer attribute sequences, where \(n=5\) in our setting.
All attribute sequences are spatially and temporally aligned with the corresponding RGB video.

Given the G-buffer conditions \(\mathbf{g}\) and a text prompt \(c\), our goal is to generate a photorealistic video that preserves the geometry and material constraints encoded by \(\mathbf{g}\), while following the global appearance specified by \(c\), such as weather, illumination, and time of day.

To map the G-buffer conditions into the spatiotemporal latent space of the pretrained video diffusion model, we independently encode each attribute sequence using the frozen video VAE encoder \(\mathcal{E}\):
\begin{equation}
z_g^{(i)}
=
\mathcal{E}\left(g^{(i)}\right),
\qquad i=1,\ldots,n.
\end{equation}
Here, \(z_g^{(i)}\) denotes the latent representation of the \(i\)-th G-buffer attribute sequence.

This maps all attributes to the same spatial and temporal resolution as the video latent while preserving their individual information.
We then concatenate the encoded G-buffer features with the noisy video latent $z_t$ along the channel dimension:
\begin{equation}
z_{\mathrm{cat}}
=
\left[
z_t
, z_g^{(1)}
, \cdots ,
 z_g^{(n)}
\right].
\end{equation}

A lightweight 3D convolutional projector $\mathcal{F}_{\mathrm{proj}}$ is used to integrate these features:
\begin{equation}
\tilde{z}_t
=
\mathcal{F}_{\mathrm{proj}}
\left(z_{\mathrm{cat}}\right).
\end{equation}
The projector captures spatial-temporal interactions and the relations among different G-buffer attributes, and maps the fused features back to the original latent channel dimension. The resulting latent $\tilde{z}_t$ is finally passed to the DiT backbone together with the timestep $t$ and text prompt $c$.

\subsection{Two-Stage Sim-to-Real Adaptation}\label{sec:training}

Real-world videos provide realistic appearance, but their estimated G-buffers often contain noise and inverse-rendering artifacts. Directly training on these noisy pairs may cause the G-buffer projector to learn estimation errors instead of reliable relationships among geometric and material attributes. We therefore separate G-buffer relation learning from real-world appearance adaptation.
We insert LoRA~\cite{hu2022lora} modules into the attention layers of the DiT, while keeping the pretrained DiT weights $W_0$ frozen throughout training. In Stage I, the G-buffer projector and the LoRA module, denoted by $\Delta W_{\mathrm{syn}}$, are jointly trained on synthetic videos with accurate G-buffers. This stage allows the projector to learn the relations among geometry and material attributes and enables the diffusion backbone to follow the fused conditions. In Stage II, we load the learned projector and $\Delta W_{\mathrm{syn}}$ into the DiT and keep both frozen. We then introduce an additional LoRA module, denoted by $\Delta W_{\mathrm{real}}$, and only train the LoRA module on real-world videos with estimated G-buffers. In this way, the model adapts to real-world lighting, textures, materials, and other appearance statistics without disturbing the G-buffer relations learned in Stage I.

For a clean video latent $z_0$, Flow Matching constructs
\begin{equation}
z_t=(1-t)z_0+t\epsilon,
\qquad
u_t=\frac{\mathrm{d}z_t}{\mathrm{d}t}=\epsilon-z_0,
\label{eq:flow_path}
\end{equation}
where $\epsilon\sim\mathcal{N}(0,I)$ and $t\sim\mathcal{U}(0,1)$.

\paragraph{\textbf{Stage I: G-buffer alignment via Synthetic Data.}}

In the first stage, our primary goal is to enable the model to comprehend the underlying physical meanings of the G-buffers and learn how to accurately map these intrinsic physical priors into the visual generation process.
Let $\mathcal{D}_{\mathrm{syn}}$ denote the high-quality synthetic dataset comprising tuples $(x^{\mathrm{syn}}, g^{\mathrm{syn}}, c^{\mathrm{syn}})$ of rendered videos, perfectly aligned G-buffers, and text prompts. During this stage, we jointly optimize the parameters of the G-buffer projector, denoted by $\theta_{\mathrm{proj}}$, and the Stage-I LoRA modules $\Delta W_{\mathrm{syn}}$. With $W_{\mathrm{I}}=W_0+\Delta W_{\mathrm{syn}}$, the training objective is
\begin{equation}
\begin{aligned}
\mathcal{L}_{\mathrm{syn}}
=\mathbb{E}\big[\|u_t^{\mathrm{syn}}-
 v_{\theta}(\tilde z_t^{\mathrm{syn}},t,c^{\mathrm{syn}};
 W_{\mathrm{I}})\|_2^2\big],
\end{aligned}
\label{eq:stage1_objective}
\end{equation}
where $z_0^{\mathrm{syn}}$ is obtained by encoding $x^{\mathrm{syn}}$ with the pretrained VAE, $z_t^{\mathrm{syn}}$ and $u_t^{\mathrm{syn}}$ follow Eq.~\eqref{eq:flow_path}, $\tilde z_t^{\mathrm{syn}}$ fuses the noisy video latent with the projected G-buffer features, and $v_\theta$ predicts the target flow velocity. By optimizing this objective, the model learns the physical and spatial correspondence between the G-buffer features and the generated visual content.

\begin{figure*}[!t]
  \centering
  \includegraphics[width=0.90\textwidth]{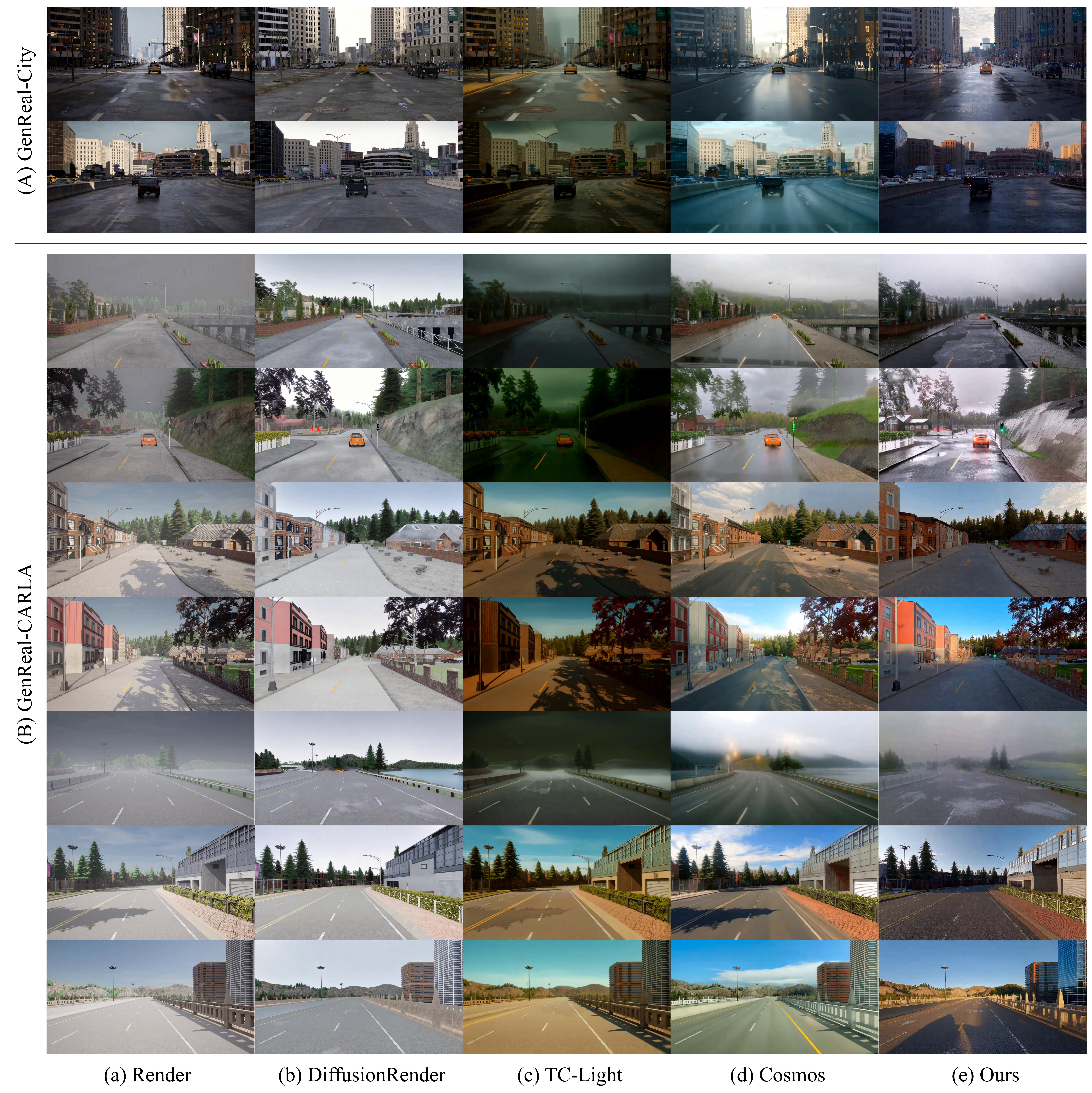}

  \caption{Qualitative results on the synthetic test sets (GenReal-City and GenReal-CARLA). Unlike baselines that retain artificial simulator aesthetics, our approach leverages the explicit geometry and physical priors of G-buffers, alongside text prompts, to achieve photorealistic sim-to-real synthesis while maintaining structural integrity.}

  \label{fig:sresult}
\end{figure*}
\input{Parts/031_compare_table}

\paragraph{\textbf{Stage II: Style Adaptation via Real-World Data.}}

While Stage I successfully aligns the generative process with the underlying geometry and material constraints, the resulting illumination and material appearance inherently retain the ``simulated'' style.
To achieve sim-to-real translation, we perform a second training stage using a real-world dataset, denoted as $\mathcal{D}_{\mathrm{real}}$, which contains real videos, their corresponding inverse-rendered G-buffers, and text prompts $(x^{\mathrm{real}}, g^{\mathrm{inv}}, c^{\mathrm{real}})$.
Prior to training, we explicitly initialize the framework with the learned weights $\theta_{\mathrm{proj}}^*$ and $\Delta W_{\mathrm{syn}}^*$ acquired in Stage I. Crucially, to prevent the noisy artifacts of inverse-rendered G-buffers from degrading the model, the projector weights are kept frozen.
With $W_{\mathrm{II}}=W_0+\Delta W_{\mathrm{syn}}^*+\Delta W_{\mathrm{real}}$, we optimize the style-adaptation objective strictly with respect to the new Stage-II LoRA modules $\Delta W_{\mathrm{real}}$:
\begin{equation}
\begin{aligned}
\mathcal{L}_{\mathrm{real}}
=\mathbb{E}\big[\|u_t^{\mathrm{real}}-
 v_{\theta}(\tilde z_t^{\mathrm{real}},t,c^{\mathrm{real}};
 W_{\mathrm{II}})\|_2^2\big],
\end{aligned}
\label{eq:stage2_objective}
\end{equation}
where $z_0^{\mathrm{real}}$ is obtained by encoding $x^{\mathrm{real}}$ with the pretrained VAE, $z_t^{\mathrm{real}}$ and $u_t^{\mathrm{real}}$ follow Eq.~\eqref{eq:flow_path}, and $\tilde z_t^{\mathrm{real}}$ fuses the noisy video latent with the projected inverse-rendered G-buffer features. During Stage II, $W_0$, $\theta_{\mathrm{proj}}^*$, and $\Delta W_{\mathrm{syn}}^*$ are kept frozen, and only $\Delta W_{\mathrm{real}}$ is optimized. Guided by the text prompt $c$, this stage efficiently shifts the model's output distribution, adapting lighting, texture appearance, and complex optical effects to real-world statistics, all while preserving the robust physical rendering logic locked in by the frozen projector $\theta_{\mathrm{proj}}^*$.%

%% file: Parts/031_compare_table.tex
\begin{table*}[t]
  \centering
  \caption{Quantitative comparison on GenReal-City and GenReal-CARLA.}
  \label{tab:quan_syn_data}

  \footnotesize
  \setlength{\tabcolsep}{4pt}
  \resizebox{\textwidth}{!}{%
    \begin{tabular}{@{}lcccccccccc@{}}
      \toprule
      \multirow{2}{*}{Method}
      & \multicolumn{4}{c}{GenReal-City}
      & \multicolumn{4}{c}{GenReal-CARLA}
      & \multicolumn{2}{c}{Efficiency} \\
      \cmidrule(lr){2-5}
      \cmidrule(lr){6-9}
      \cmidrule(lr){10-11}

      & FVD$\downarrow$
      & CLIP-T$\uparrow$
      & NDE$\downarrow$
      & VLM-BT$\uparrow$
      & FVD$\downarrow$
      & CLIP-T$\uparrow$
      & NDE$\downarrow$
      & VLM-BT$\uparrow$
      & VRAM (GB)$\downarrow$
      & FPS$\uparrow$ \\
      \midrule

      DiffRender~\cite{DiffusionRenderer}
      & 1116.45 & 0.2836 & 0.0260 & -0.1960
      & 1113.35 & 0.2819 & \textbf{0.0302} & -0.2180
      & 94.55 & \underline{0.348} \\

      TC-Light~\cite{liu2025tclight}
      & 1117.74 & 0.2953 & \textbf{0.0238} & -0.0752
      & 984.37 & 0.2878 & \underline{0.0316} & -0.1737
      & \underline{19.82} & 0.288 \\

      Cosmos~\cite{cosmos_cookbook_control_modalities_2025}
      & \underline{997.26} & \underline{0.3006}
      & 0.0251 & \underline{-0.0070}
      & \underline{981.75} & \textbf{0.2951}
      & 0.0335 & \underline{0.1571}
      & 57.74 & 0.142 \\

      \midrule

      Ours
      & \textbf{979.69} & \textbf{0.3007}
      & \underline{0.0243} & \textbf{0.2783}
      & \textbf{599.71} & \underline{0.2949}
      & 0.0324 & \textbf{0.2346}
      & \textbf{13.67} & \textbf{0.598} \\

      \bottomrule
    \end{tabular}%
  }

\end{table*}

%% file: Parts/05_Experiments.tex
\section{Experiments}
\label{sec:experiment}

\subsection{Experimental Setup}

\begin{figure*}[!t]
  \centering
  % Crop only the embedded label strip; typeset readable labels below.
  \includegraphics[width=\textwidth,trim=0 108bp 0 0,clip]{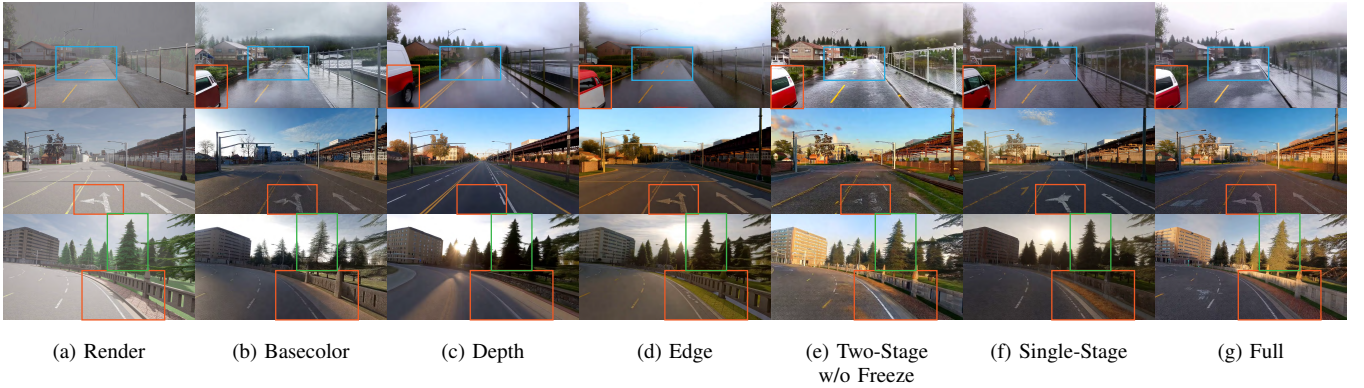}\par
  \vspace{2pt}
  \noindent\panellabel{(a) Render}%
  \panellabel{(b) Basecolor}%
  \panellabel{(c) Depth}%
  \panellabel{(d) Edge}%
  \panellabel{(e) Two-Stage\\w/o Freeze}%
  \panellabel{(f) Single-Stage}%
  \panellabel{(g) Full}\par

  \caption{Qualitative comparison of control signals and training strategies. Depth and edge controls produce overly smooth results with semantic, texture, color, and lighting errors. Basecolor control provides richer appearance cues but may flatten geometric structures. Two-Stage w/o Freeze degrades learned lighting and shadow patterns, while Single-Stage training may distort geometry and road markings. Our full model best preserves geometry, markings, textures, lighting, and shadows.}
  \label{fig:abla_result}

\end{figure*}

\paragraph{\textbf{Implementation Details.}}

We build NeuralSRender on the pretrained Wan2.2-TI2V-5B model~\cite{wan2025wan}. We use AdamW with a weight decay of $1\times10^{-4}$ and set the LoRA rank to 32 in both stages. The pretrained VAE and DiT weights remain frozen throughout training.
In Stage I, we jointly train the G-buffer projector $\mathcal{F}_{\mathrm{proj}}$ and the Stage-I LoRA parameters $\Delta W_{\mathrm{syn}}$ on GenReal-City for 6 epochs. In Stage II, we freeze $\mathcal{F}_{\mathrm{proj}}$ and train a newly initialized set of LoRA parameters $\Delta W_{\mathrm{real}}$ on GenReal-DV for 2 epochs. The batch size is 4. The whole training takes approximately five days on four RTX 4090 GPUs.

\paragraph{Metrics.}
For generation quality, we compute FVD against the GenReal-DV reference set, use LongCLIP~\cite{zhang2024longclip} for text--video alignment (CLIP-T), and report VLM-BT for frame-level photorealism. For VLM-BT, Gemini 3 Flash, GPT-5.4 mini, and Qwen3.6-Plus rank randomly shuffled outputs, and their pairwise preferences are aggregated using a Bradley--Terry model. Since the Bradley--Terry model is fitted independently for each comparison set, VLM-BT scores reflect relative preferences within that set and may vary for the same method when the compared methods change. For geometric consistency, we estimate relative depth from generated and source videos using Video Depth Anything~\cite{video_depth_anything}, and report the mean absolute normalized difference as NDE. For efficiency, we measure peak VRAM and FPS at a resolution of $1280\times704\times81$  on a single NVIDIA RTX Pro 6000 GPU.

\paragraph{Baselines.}
We compare NeuralSRender with DiffusionRenderer~\cite{DiffusionRenderer}, TC-Light~\cite{liu2025tclight}, and Cosmos-Transfer 2.5 (Cosmos)~\cite{cosmos_cookbook_control_modalities_2025}. For DiffusionRenderer, we use the same G-buffers as our method and estimate environment illumination maps with DiffusionLight-Turbo~\cite{Chinchuthakun2025DiffusionLightTurbo}, following its official relighting pipeline. For TC-Light, we use the official pretrained weights and default settings. For Cosmos, we use the edge-controlled variant with edge maps extracted from the source videos. All methods use the same test clips and output resolution. Identical prompts are provided to all text-conditioned methods, while DiffusionRenderer is evaluated without text conditioning.

\subsection{Comparison Results}

Table~\ref{tab:quan_syn_data} reports the quantitative results on GenReal-City and GenReal-CARLA. Ours achieves the best FVD and VLM-BT scores on both test sets, demonstrating improved distributional similarity and perceived photorealism. Meanwhile, it maintains competitive CLIP-T scores and NDE. Overall, our method improves visual realism without substantially sacrificing text alignment or geometric consistency. Qualitative comparisons in Fig.~\ref{fig:sresult} further show that NeuralSRender achieves a favorable balance between visual realism and structural consistency.
DiffusionRenderer ~\cite{DiffusionRenderer} closely follows the geometry and material cues provided by the G-buffers. However, its outputs often retain the artificial appearance of simulator renderings, with limited improvements in textures and illumination.
TC-Light~ \cite{liu2025tclight} takes the rendered RGB video as input and mainly adjusts global color tone. As shown in Fig.~\ref{fig:sresult}(c), the original textures and cast shadows often remain unchanged, which can conflict with the modified appearance.
Cosmos~\cite{cosmos_cookbook_control_modalities_2025} produces visually appealing results using strong generative priors. However, edge-based conditioning provides only sparse structural guidance. Edges extracted from rendered videos may also contain shadow boundaries and noisy details, leading to geometric deviations, artificial lens flares, or spurious objects, as illustrated in Fig.~\ref{fig:sresult}(d). Edge maps contain no explicit material information. This limitation often results in overly smooth surfaces, unrealistic reflections, and limited texture detail, making the outputs less photorealistic.
In contrast, NeuralSRender employs dense G-buffers to provide geometry and material guidance, while text prompts control global attributes such as weather and illumination.
This combination produces realistic reflections and rough-surface appearances while preserving scene structure and reducing artifacts, as shown in Fig.~\ref{fig:sresult}(e).

\subsection{Ablation Studies}

\input{Parts/032_abla_table}

\begin{figure*}[!t]
  \centering
  % Crop only the embedded label strip; preserve the full image grid.
  \includegraphics[width=\textwidth,trim=0 80bp 0 0,clip]{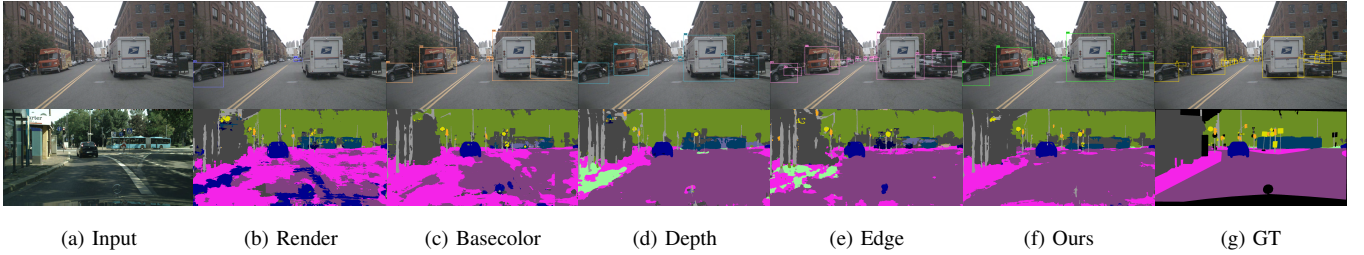}\par
  \vspace{2pt}
  \noindent\panellabel{(a) Input}%
  \panellabel{(b) Render}%
  \panellabel{(c) Basecolor}%
  \panellabel{(d) Depth}%
  \panellabel{(e) Edge}%
  \panellabel{(f) Ours}%
  \panellabel{(g) GT}\par
  \caption{Qualitative perception results with different synthetic training data.}
  \label{fig:downstream_result}
\end{figure*}

% \begin{figure*}[!t]
%   \centering
%   \includegraphics[width=\textwidth]{Figures/long_video_4017_merge.jpg}
%   \caption{Qualitative comparison of long video generation. Our model can generate long video up to 4,100 frames while keep geometry and appearance consistency.}
%   \label{fig:longvideo_result}
% \end{figure*}

We ablate different control signals and training strategies, including Basecolor Control, Depth Control, Edge Control, Two-Stage w/o Freeze, and Single-Stage Training. The first three variants use only the corresponding control signal. Two-Stage w/o Freeze retains the two-stage synthetic-to-real training procedure but keeps the G-buffer projector trainable during real-world adaptation.
Single-Stage Training mixes synthetic and real-world samples and jointly trains the model on both domains in a single stage.  All experiments use the same data and number of training steps.

As shown in Fig.~\ref{fig:abla_result} and Table~\ref{tab:ablation},
\textbf{Depth Control} has difficulty representing both near and distant regions in outdoor scenes. It also cannot constrain road markings or scene colors.
\textbf{Edge Control} provides structural guidance, but it contains no color or texture information. As a result, both controls provide incomplete geometric and appearance constraints, often leading to semantic errors, overly smooth surfaces, limited texture detail, and unrealistic lighting or shadows.
\textbf{Basecolor Control} provides richer appearance information but offers limited cues for distinguishing geometry and material properties. Geometrically distinct regions may share similar basecolors, causing the model to merge them into a flat surface. For example, the curb in the orange box of Fig.~\ref{fig:abla_result} loses its clear geometric structure, reducing geometric accuracy.
In contrast, full G-buffer conditioning combines complementary geometry and material cues, enabling better structural preservation and more realistic appearance.

For the training strategies under full G-buffer conditioning, \textbf{Two-Stage w/o Freeze} is strongly affected by the noisy real-data G-buffers in the second training stage. Although it can preserve the overall scene geometry, the previously learned relations are partially destroyed, leading to a loss of realistic lighting and shadow patterns.
\textbf{Single-stage} training produces results closest to those of the full model. Nevertheless, the noisy real data can still interfere with the geometry and semantic structures learned from synthetic data. For example, the rear of a car may be incorrectly generated as its front, and ground markings may be distorted (orange box in Fig.~\ref{fig:abla_result}). Finally, our full model freezes the G-buffer projector during real-world adaptation, preserving the learned attribute relations while LoRA adapts the model to real-world appearance.
It achieves the best FVD, NDE, and VLM-BT scores, with competitive CLIP-T scores and strong geometric and visual quality.

\subsection{Sim-to-Real Transfer for Autonomous Driving}

We evaluate sim-to-real transfer for object detection and semantic segmentation without target-domain fine-tuning.
For each setting, we train perception models on either raw rendered videos or videos generated with different control signals from the same CARLA scenes. The trained models are then directly evaluated on unseen real-world datasets without target-domain fine-tuning.
For object detection, we use YOLOv8~\cite{varghese2024yolov8} and evaluate on nuImages~\cite{nuscenes2019}. As reported in Table~\ref{tab:downstream}, our method achieves the highest detection AP and AP$_{50}$ among the evaluated settings, increasing AP$_{50}$ from 35.1 to 40.6 compared with training on the original CARLA renders. For semantic segmentation, we use SegFormer~\cite{xie2021segformer} and evaluate on Cityscapes~\cite{cordts2016cityscapes}. Our method again performs best, improving mIoU from 32.4 to 40.8  over the CARLA renders. Fig.~\ref{fig:downstream_result} complements these quantitative
results with qualitative comparisons: models trained on our
translated data detect more of the annotated vehicles and produce
a more coherent road segmentation than those trained on the
original CARLA renders in the illustrated examples.
These gains demonstrate the effectiveness of the translated data for training perception models that generalize to real-world driving scenes.

% NeuralSRender also supports long-video generation of up to 4,100 frames. Please refer to the multimedia materials.
\input{Parts/033_downstream_table}

%% file: Parts/032_abla_table.tex
\begin{table}[t]

  \centering
  \caption{Quantitative results of the ablation study.}
    \label{tab:ablation}

  \footnotesize
  \renewcommand{\arraystretch}{1.05}
  \setlength{\tabcolsep}{3pt}

  \begin{tabular*}{\linewidth}{
    @{\extracolsep{\fill}}lcccc@{}
  }
    \toprule
    Method
    & FVD$\downarrow$
    & CLIP-T$\uparrow$
    & NDE$\downarrow$
    & VLM-BT$\uparrow$ \\
    \midrule

    (a) Basecolor Only
    & 669.41
    & 0.2925
    & 0.0410
    & 0.0557 \\

    (b) Depth Only
    & 812.12
    & \textbf{0.2968}
    & 0.0602
    & -0.1499 \\

    (c) Edge Only
    & 775.24
    & 0.2943
    & 0.0401
    & -0.0986 \\

    \midrule

    (d) Single-Stage
    & \underline{653.45}
    & 0.2945
    & \underline{0.0347}
    & \underline{0.1088} \\

    (e) Two-Stage w/o Freeze
    & 730.49
    & \underline{0.2957}
    & 0.0387
    & -0.0328 \\

    (f) Full Model
    & \textbf{599.71}
    & 0.2949
    & \textbf{0.0324}
    & \textbf{0.1168} \\

    \bottomrule
  \end{tabular*}

\end{table}

%% file: Parts/033_downstream_table.tex
\begin{table}[t]

  \centering
\caption{Perception performance with different synthetic training data.}
  \label{tab:downstream}

  \footnotesize
  \begin{tabular*}{\linewidth}{@{\extracolsep{\fill}}lcccc@{}}
    \toprule
    Method
    & AP$\uparrow$
    & AP$_{50}\uparrow$
    & mIoU$\uparrow$
    & mAcc$\uparrow$ \\
    \midrule
    (a) Render
    & 16.7 & 35.1 & 32.4 & 42.3 \\
    (b) Basecolor Only
    & 18.8 & 38.1 & 37.6 & 45.6 \\
    (c) Depth Only
    & 14.7 & 30.9 & 37.4 & 47.3 \\
    (d) Edge Only
    & 17.1 & 37.4 & 35.2 & 45.9 \\
    (e) Ours
    & \textbf{19.4}
    & \textbf{40.6}
    & \textbf{40.8}
    & \textbf{50.2} \\
    \bottomrule
  \end{tabular*}

\end{table}

%% file: Parts/06_Conclusion.tex
\section{Conclusion}

We present NeuralSRender, a parameter-efficient framework for G-buffer-guided sim-to-real video translation.
Its two-stage training separates reliable conditioning learning from real-world appearance adaptation.
We also introduce GenReal, a driving video dataset containing synthetic and real-world videos with corresponding G-buffers.
Experiments demonstrate improved visual realism with competitive geometric consistency.
Overall, our work enables the generation of high-quality, scalable video data for autonomous driving models, highlighting the broader potential of sim-to-real video translation for autonomous driving.

\label{sec:conlusion}